# The Deep Learning Revolution and Its Implications for Computer Architecture and Chip Design

*Jeffrey Dean*
*Google Research*
jeff@google.com

## Abstract

The past decade has seen a remarkable series of advances in machine learning, and in particular deep learning approaches based on artificial neural networks, to improve our abilities to build more accurate systems across a broad range of areas, including computer vision, speech recognition, language translation, and natural language understanding tasks. This paper is a companion paper to a keynote talk at the 2020 International Solid-State Circuits Conference (ISSCC) discussing some of the advances in machine learning, and their implications on the kinds of computational devices we need to build, especially in the post-Moore's Law-era. It also discusses some of the ways that machine learning may also be able to help with some aspects of the circuit design process. Finally, it provides a sketch of at least one interesting direction towards much larger-scale multi-task models that are sparsely activated and employ much more dynamic, example- and task-based routing than the machine learning models of today.

## Introduction

The past decade has seen a remarkable series of advances in machine learning (ML), and in particular deep learning approaches based on artificial neural networks, to improve our abilities to build more accurate systems across a broad range of areas [LeCun *et al.* 2015]. Major areas of significant advances include computer vision [Krizhevsky *et al.* 2012, Szegedy *et al.* 2015, He et al. 2016, Real *et al.* 2017, Tan and Le 2019], speech recognition [Hinton *et al.* 2012, Chan *et al.* 2016], language translation [Wu *et al.* 2016] and other natural language tasks [Collobert *et al.* 2011, Mikolov et al. 2013, Sutskever *et al.* 2014, Shazeer *et al.* 2017, Vaswani et al. 2017, Devlin *et al.* 2018]. The machine learning research community has also been able to train systems to accomplish some challenging tasks by learning from interacting with environments, often using reinforcement learning, showing success and promising advances in areas such as playing the game of Go [Silver *et al.* 2017], playing video games such as Atari games [Mnih *et al.* 2013, Mnih *et al.* 2015] and Starcraft [Vinyals *et al.* 2019], accomplishing robotics tasks such as substantially improved grasping for unseen objects [Levine *et al.* 2016, Kalashnikov *et al.* 2018], emulating observed human behavior [Sermanet *et al.* 2018], and navigating complex urban environments using autonomous vehicles [Angelova *et al.* 2015, Bansal *et al.* 2018].

As an illustration of the dramatic progress in the field of computer vision, Figure 1 shows a graph of the improvement over time for the Imagenet challenge, an annual contest run by Stanford University [Deng *et al.* 2009] where contestants are given a training set of one million color images across 1000 categories, and then use this data to train a model to generalize to an evaluation set of images across the same categories. In 2010 and 2011, prior to the use of deep learning approaches in this contest, the winning entrants used hand-engineered computer vision features and the top-5 error rate was above 25%. In 2012, Alex Krishevsky, Ilya Sutskever, and Geoffrey Hinton used a deep neural network, commonly referred to as “AlexNet”, to take first place in the contest with a major reduction in the top-5 error rate to 16% [Krishevsky *et al.* 2012]. Their team was the only team that used a neural network in 2012. The next year, the deep learning computer vision revolution was in full force with the vast majority of entries from teams using deep neural networks, and the winning error rate again dropped substantially to 11.7%. We know from a careful study that Andrej Karpathy performed that human error on this task is just above 5% if the human practices for ~20 hours, or 12% if a different person practices for just a few hours [Karpathy 2014]. Over the course of the years 2011 to 2017, the winning Imagenet error rate dropped sharply from 26% in 2011 to 2.3% in 2017.

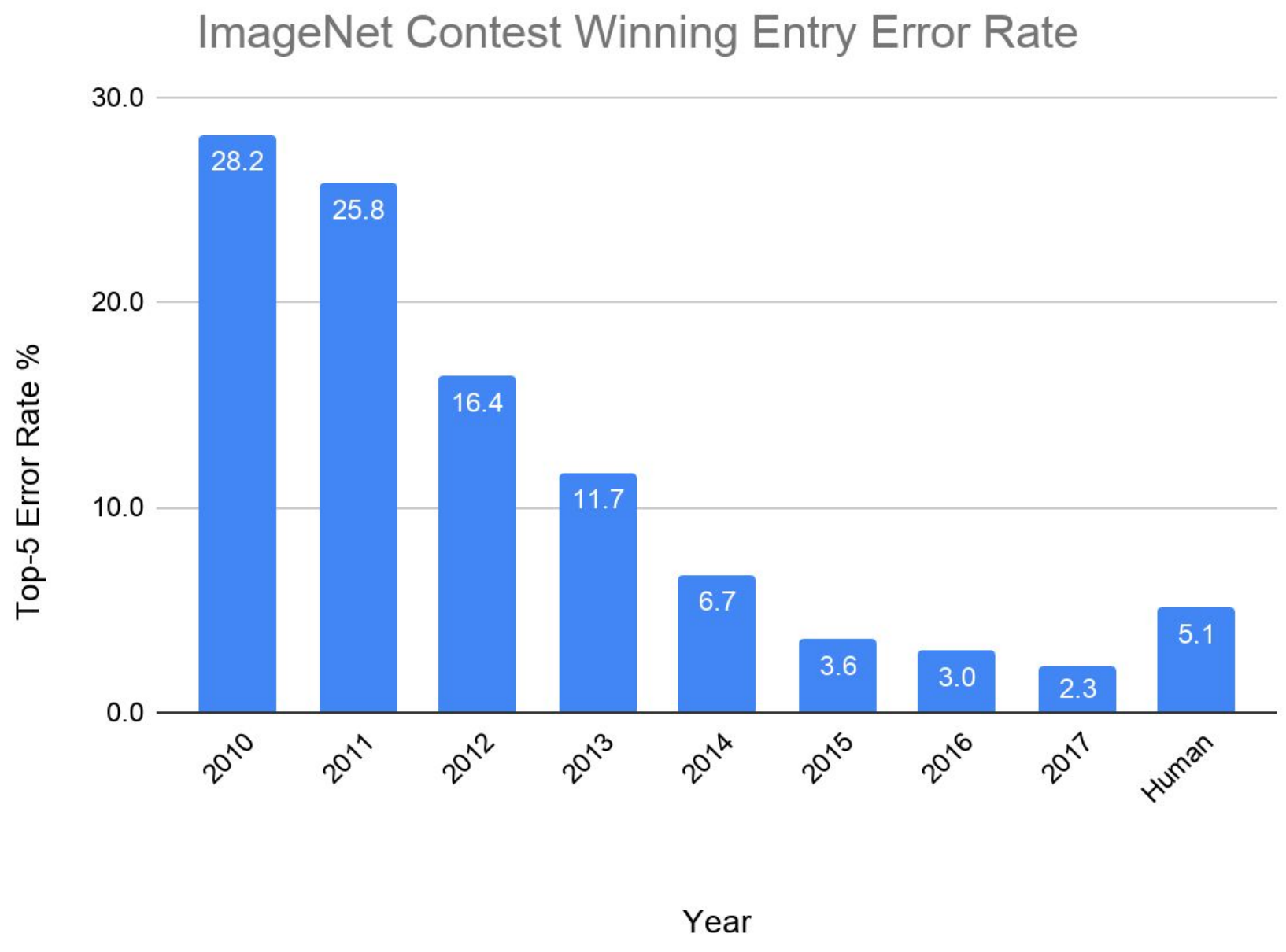

Figure 1: ImageNet classification contest winner accuracy over time

These advances in fundamental areas like computer vision, speech recognition, language understanding, and large-scale reinforcement learning have dramatic implications for many fields. We have seen a steady series of results in many different fields of science and medicine by applying the basic research results that have been generated over the past decade to these problem areas. Examples include promising areas of medical imaging diagnostic tasks including for diabetic retinopathy [Gulshan *et al.* 2016, Krause *et al.* 2018], breast cancer pathology [Liu *et al.* 2017], lung cancer CT scan interpretation [Ardila *et al.* 2019], and dermatology [Esteva *et al.* 2017]. Sequential prediction methods that are useful for language translation also turn out to be useful for making accurate predictions for a variety of different medically-relevant tasks from electronic medical records [Rajkomar *et al.* 2018]. These early signs point the way for machine learning to have a significant impact across many areas of health and medical care [Rajkomar *et al.* 2019, Esteva *et al.* 2019].

Other fields that have been improved by the use of deep learning-based approaches include quantum chemistry [Gilmer *et al.* 2017], earthquake prediction [DeVries *et al.* 2018], flood forecasting [Nevo 2019], genomics [Poplin *et al.* 2018], protein folding [Evans *et al.* 2018], high energy physics [Baldi *et al.* 2014], and agriculture [Ramcharan *et al.* 2017].

With these significant advances, it is clear that the potential for ML to change many different fields of endeavor is substantial.

# Moore's Law, Post Moore's Law, and the Computational Demands of Machine Learning

Many of the key ideas and algorithms underlying deep learning and artificial neural networks have been around since the 1960s, 1970s, 1980s, and 1990s [Minsky and Papert 1969, Rumelhart *et al.* 1988, Tesauro 1994]. In the late 1980s and early 1990s there was a surge of excitement in the ML and AI community as people realized that neural networks could solve some problems in interesting ways, with substantial advantages stemming from their ability to accept very raw forms of (sometimes heterogeneous) input data and to have the model automatically build up hierarchical representations in the course of training the model to perform some predictive task. At that time, though, computers were not powerful enough to allow this approach to work on anything but small, almost toy-sized problems. Some work at the time attempted to extend the amount of computation available for training neural networks by using parallel algorithms [Shaw 1981, Dean 1990], but for the most part, the focus of most people in the AI and ML community shifted away from neural network-based approaches. It was not until the later parts of the decade of the 2000s, after two more decades of computational performance improvements driven by Moore's Law that computers finally started to become powerful enough to train large neural networks on realistic, real-world problems like Imagenet [*Deng et al. 2009*], rather than smaller-scale, toy problems like MNIST [LeCun *et al.* 2000] and CIFAR [Krizhevsky *et al.* 2009]. In particular, the paradigm of general-purpose computing on GPU cards (GPGPU) [Luebke *et al.* 2006], because of GPU cards' high floating point performance relative to CPUs, started to allow neural networks to show interesting results on difficult problems of real consequence.

It is perhaps unfortunate that just as we started to have enough computational performance to start to tackle interesting real-world problems and the increased scale and applicability of machine learning has led to a dramatic thirst for additional computational resources to tackle larger problems, the computing industry as a whole has experienced a dramatic slowdown in the year-over-year improvement of general purpose CPU performance. Figure 2 shows this dramatic slowdown, where we have gone from doubling general-purpose CPU performance every 1.5 years (1985 through 2003) or 2 years (2003 to 2010) to now being in an era where general purpose CPU performance is expected to double only every 20 years [Hennessy and Patterson 2017]. Figure 3 shows the dramatic surge in computational demands for some important recent machine learning advances (note the logarithmic Y-axis, with the best-fit line showing a doubling time in computational demand of 3.43 months for this select set of important ML research results) [OpenAI 2018]. Figure 4 shows the dramatic surge in research output in the field of machine learning and its applications, measured via the number of papers posted to the machine-learning-related categories of Arxiv, a popular paper preprint hosting service, with more than 32 times as many papers posted in 2018 as in 2009 (a growth rate of more than doubling every 2 years). There are now more than 100 research papers per day posted to Arxiv in the machine-learning-related subtopic areas, and this growth shows no signs of slowing down.

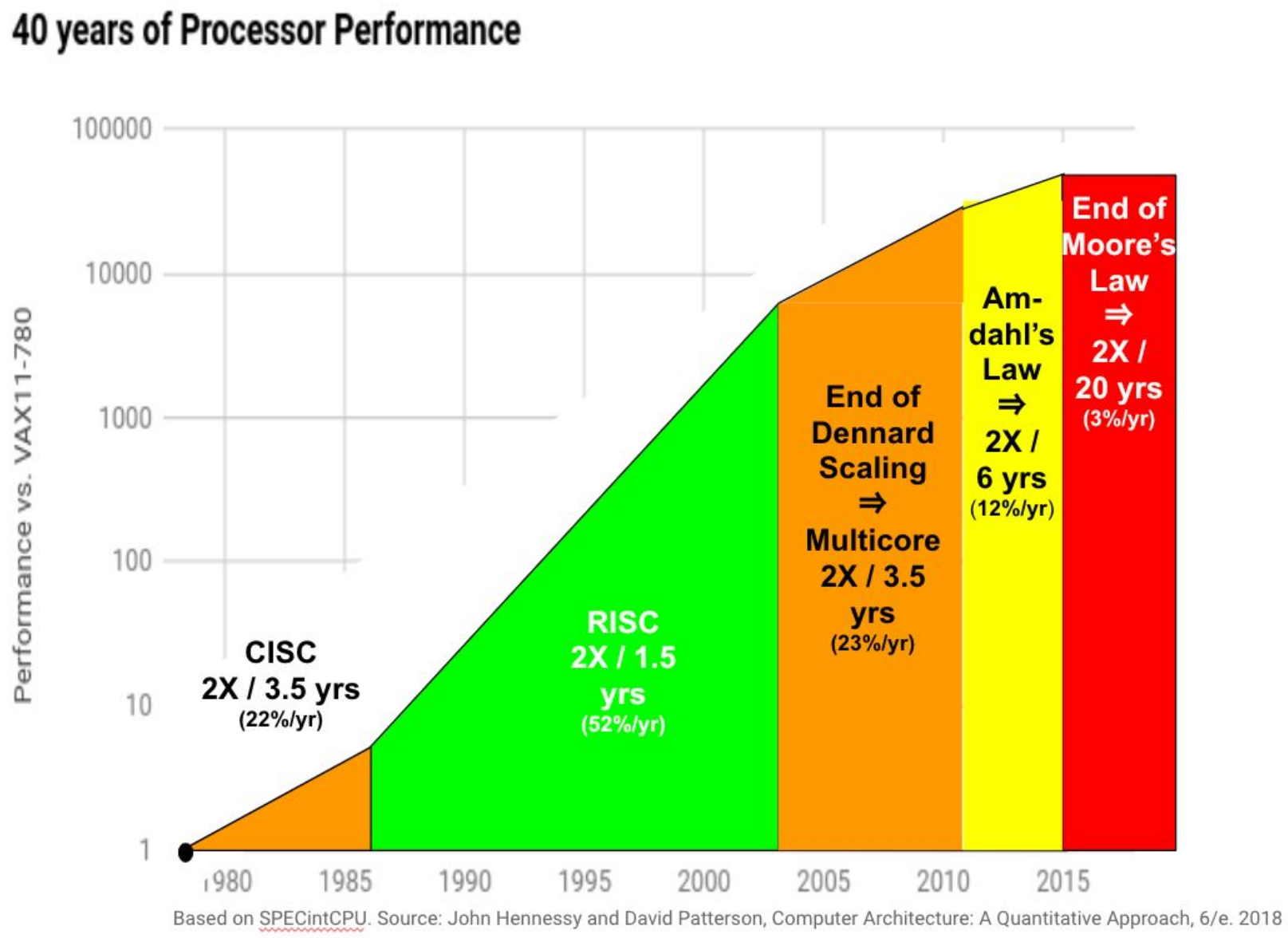

Figure 2: Computing Performance in the Moore's Law and the Post-Moore's Law Periods

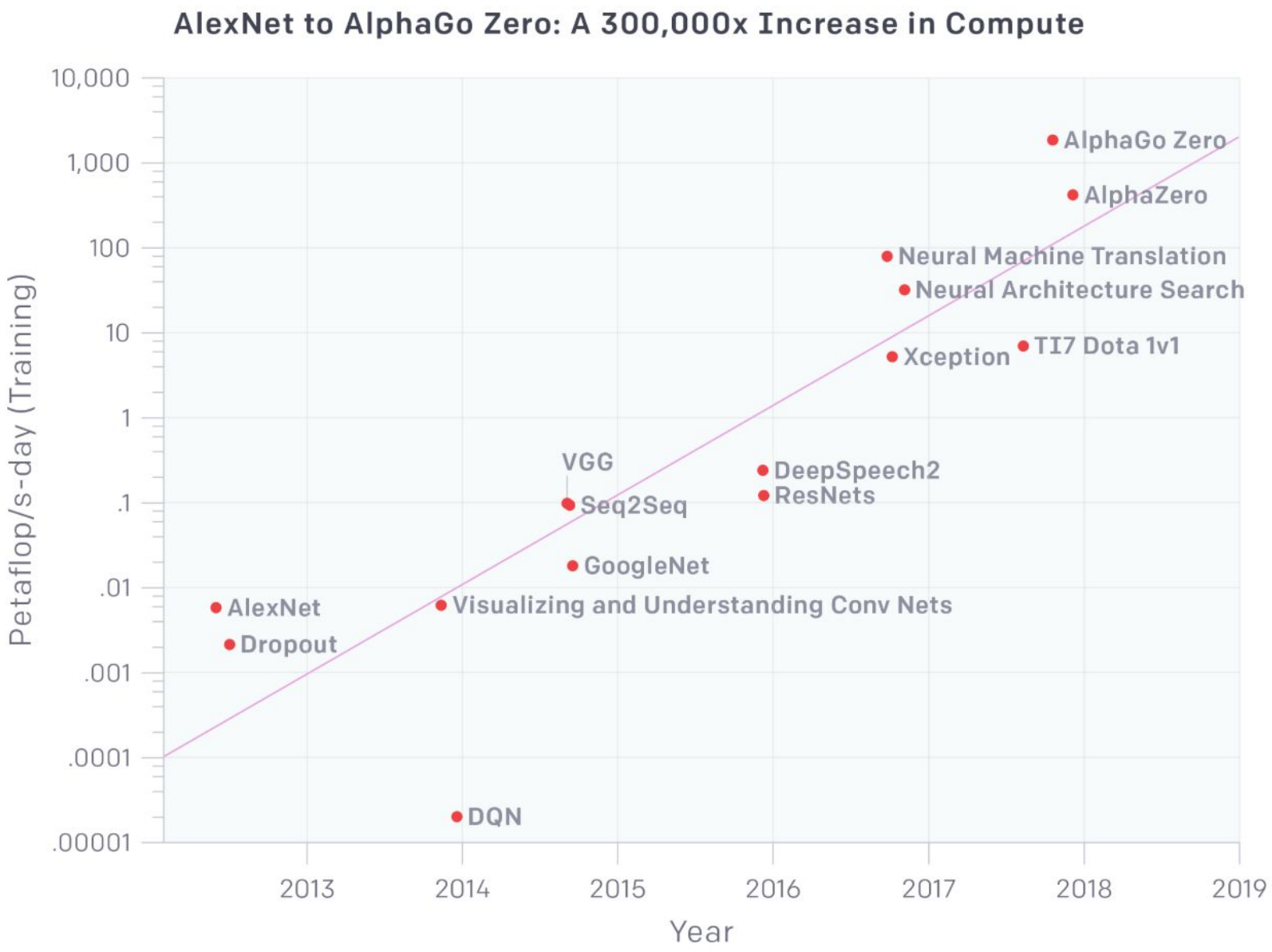

Figure 3: Some important AI Advances and their Computational Requirements
(Source: openai.com/blog/ai-and-compute/)

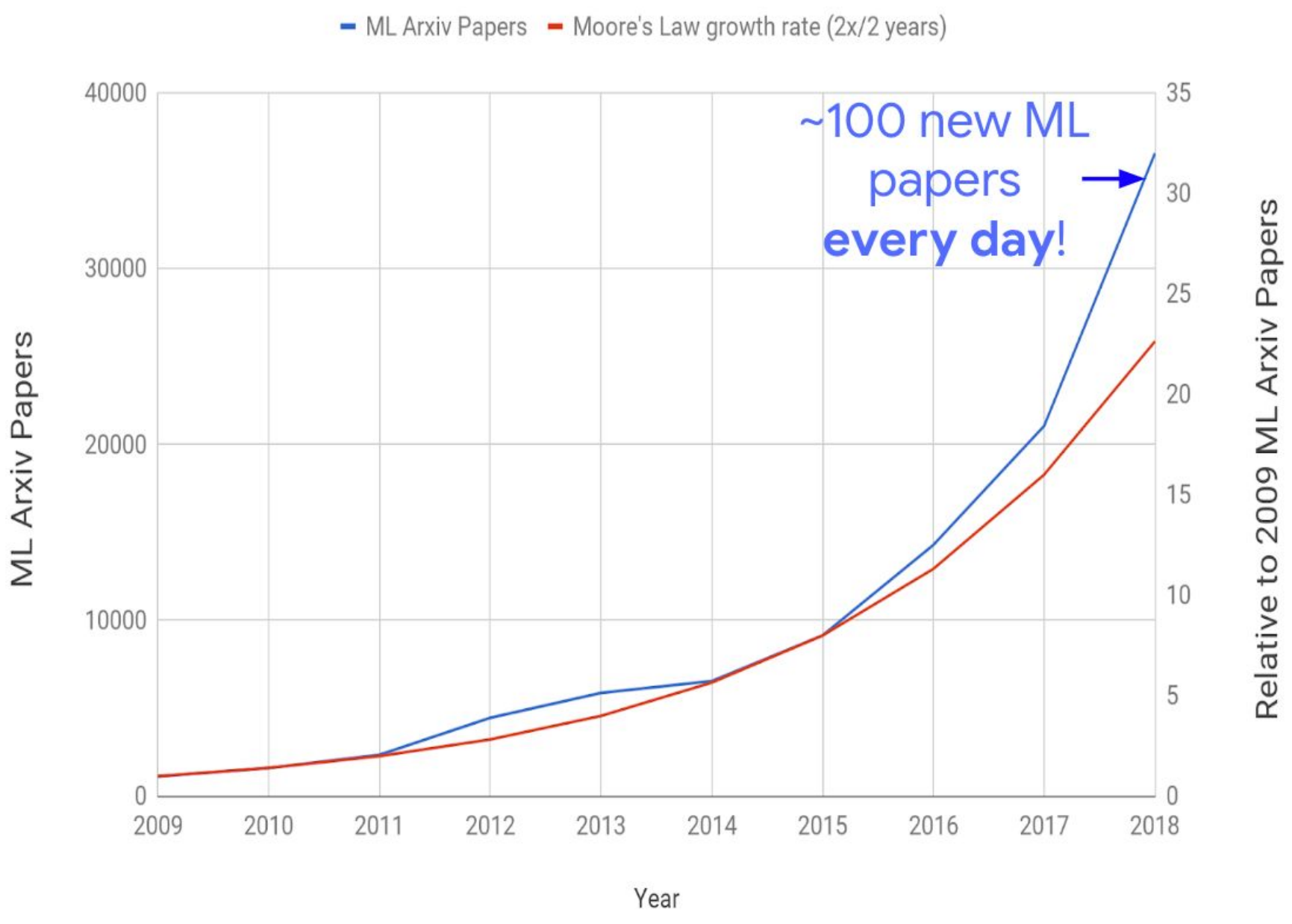

Figure 4: Machine learning-related Arxiv papers since 2009

# Machine-Learning-Specialized Hardware

In 2011 and 2012, a small team of researchers and system engineers at Google built an early distributed system called DistBelief to enable parallel, distributed training of very large scale neural networks, using a combination of model and data parallel training and asynchronous updates to the parameters of the model by many different computational replicas [Dean *et al.* 2012]. This enabled us to train much larger neural networks on substantially larger data sets and, by mid-2012, using DistBelief as an underlying framework, we were seeing dramatically better accuracy for speech recognition [Hinton *et al.* 2012] and image classification models [Le *et al.* 2012]. The serving of these models in demanding settings of systems with hundreds of millions of users, though, was another matter, as the computational demands were very large. One back of the envelope calculation showed that in order to deploy the deep neural network system that was showing significant word error rate improvements for our main speech recognition system using CPU-based computational devices would require doubling the number of computers in Google datacenters (with some bold-but-still-plausible assumptions about significantly increased usage due to more accuracy). Even if this was economically reasonable, it would still take significant time, as it would involve pouring concrete, striking arrangements for windmill farm contracts, ordering and installing lots of computers, etc., and the speech system was just the tip of the iceberg in terms of what we saw as the potential set of the application of neural networks to many of our core problems and products. This thought exercise started to get us thinking about building specialized hardware for neural networks, first for inference, and then later systems for both training and inference.

## Why Does Specialized Hardware Make Sense for Deep Learning Models?

Deep learning models have three properties that make them different than many other kinds of more general purpose computations. First, they are very tolerant of reduced-precision computations. Second, the computations performed by most models are simply different compositions of a relatively small handful of operations like matrix multiplies, vector operations, application of convolutional kernels, and other dense linear algebra calculations [Vanhoucke *et al.* 2011]. Third, many of the mechanisms developed over the past 40 years to enable general-purpose programs to run with high performance on modern CPUs, such as branch predictors, speculative execution, hyperthreaded-execution processing cores, and deep cache memory hierarchies and TLB subsystems are unnecessary for machine learning computations. So, the opportunity exists to build computational hardware that is specialized for dense, low-precision linear algebra, and not much else, but is still programmable at the level of specifying programs as different compositions of mostly linear algebra-style operations. This confluence of characteristics is not dissimilar from the observations that led to the development of specialized digital signal processors (DSPs) for telecom applications starting in the 1980s [en.wikipedia.org/wiki/Digital_signal_processor]. A key difference though, is because of the broad applicability of deep learning to huge swaths of computational problems across many domains and fields of endeavor, this hardware, despite its narrow set of supported operations, can be used for a wide variety of important computations, rather than the more narrowly tailored uses of DSPs. Based on our thought experiment about the dramatically increased computational demands of deep neural networks for some of our high volume inference applications like speech recognition and image classification, we decided to start an effort to design a series of accelerators called Tensor Processing Units for accelerating deep learning inference and training. The first such system, called TPUv1, was a single chip design designed to target inference acceleration [Jouppi *et al.* 2017].

For inference (after a model has been trained, and we want to apply the already-trained model to new inputs in order to make predictions), 8-bit integer-only calculations have been shown to be sufficient for many important models [Jouppi *et al.* 2017], with further widespread work going on in the research community to push this boundary further using things like even lower precision weights, and techniques to encourage sparsity of weights and/or activations.

The heart of the TPUv1 is a 65,536 8-bit multiply-accumulate matrix multiply unit that offers a peak throughput of 92 TeraOps/second (TOPS). TPUv1 is on average about 15X -- 30X faster than its contemporary GPU or CPU, with TOPS/Watt about 30X -- 80X higher, and was able to run production neural net applications representing about 95% of Google datacenters' neural network inference demand at the time with significant cost and power advantages [Jouppi *et al.* 2017].

Inference on low-power mobile devices is also incredibly important for many uses of machine learning. Being able to run machine learning models on-device, where the devices themselves are often the source of the raw data inputs used for models in areas like speech or vision, can have substantial latency as well as privacy benefits. It is possible to take the same design principles used for TPUv1 (a simple design targeting low precision linear algebra computations at high performance/Watt) and apply these principles to much lower power environments, such as mobile phones. Google's Edge TPU is one example of such a system, offering 4 TOps in a 2W power envelope [cloud.google.com/edge-tpu/, coral.withgoogle.com/products/]. On-device computation is already critical to many interesting use cases of deep learning, where we want computer vision, speech and other kinds of models that can run directly on sensory inputs without requiring connectivity. One such example is on-device agriculture applications, like identification of diseases in plants such as cassava, in the middle of cassava fields which may not have reliable network connectivity [Ramcharan *et al.* 2017].

With the widespread adoption of machine learning and its growing importance as a key type of computation in the world, a Cambrian-style explosion of new and interesting accelerators for machine learning computations is underway. There are more than XX venture-backed startup companies, as well as a variety of large, established companies, that are each producing various new chips and systems for machine learning. Some, such as Cerebras [www.cerebras.net/], Graphcore [www.graphcore.ai/], and Nervana (acquired by Intel) [www.intel.ai/nervana-nnp/] are focused on a variety of designs for ML training. Others, such as Alibaba [www.alibabacloud.com/blog/alibaba-unveils-ai-chip-to-enhance-cloud-computing-power_595409] are designing chips focused on inference.. Some of the designs eschew larger memory-capacity DRAM or HBM to focus on very high performance designs for models that are small enough that their entire set of parameters and intermediate values fit in SRAM. Others focus on designs that include DRAM or HBM that make them suitable for larger-scale models. Some, like Cerebras, are exploring full wafer-scale integration. Others, such as Google's Edge TPUs [cloud.google.com/edge-tpu/] are building very low power chips for inference in environments such as mobile phones and distributed sensing devices.

Designing customized machine learning hardware for training (rather than just inference) is a more complex endeavor than single chip inference accelerators. The reason is that single-chip systems for training are unable to solve many problems that we want to solve in reasonable periods of time (e.g. hours or days, rather than weeks or months), because a single-chip system cannot deliver sufficient computational power. Furthermore, the desire to train larger models on larger data sets is such that, even if a single chip could deliver enough computation to solve a given problem in a reasonable amount of time, that would just mean that we would often want to solve even larger problems (necessitating the use of multiple chips in a parallel or distributed system anyway). Therefore, designing training systems is really about designing larger-scale, holistic computer systems, and requires thinking about individual accelerator chip design, as well as high performance interconnects to form tightly coupled machine learning supercomputers. Google's second- and third-generation TPUs, TPUv2 and TPUv3 [cloud.google.com/tpu/], are designed to support both training and inference, and the basic individual devices, each consisting of four chips, were designed to be connected together into larger configurations called pods. Figure 5 shows the block diagram of a single Google TPUv2 chip, with two cores, with the main computational capacity in each core provided by a large matrix multiply unit that can yield the results of multiplying a pair of 128x128 matrices each cycle. Each chip has 16 GB (TPUv2) or 32 GB (TPUv3) of attached high-bandwidth memory (HBM). Figure 6 shows the deployment form of a Google's TPUv3 Pod of 1024 accelerator chips, consisting of eight racks of chips and accompanying servers, with the chips connected together in a 32x32 toroidal mesh, providing a peak system performance of more than 100 petaflop/s.

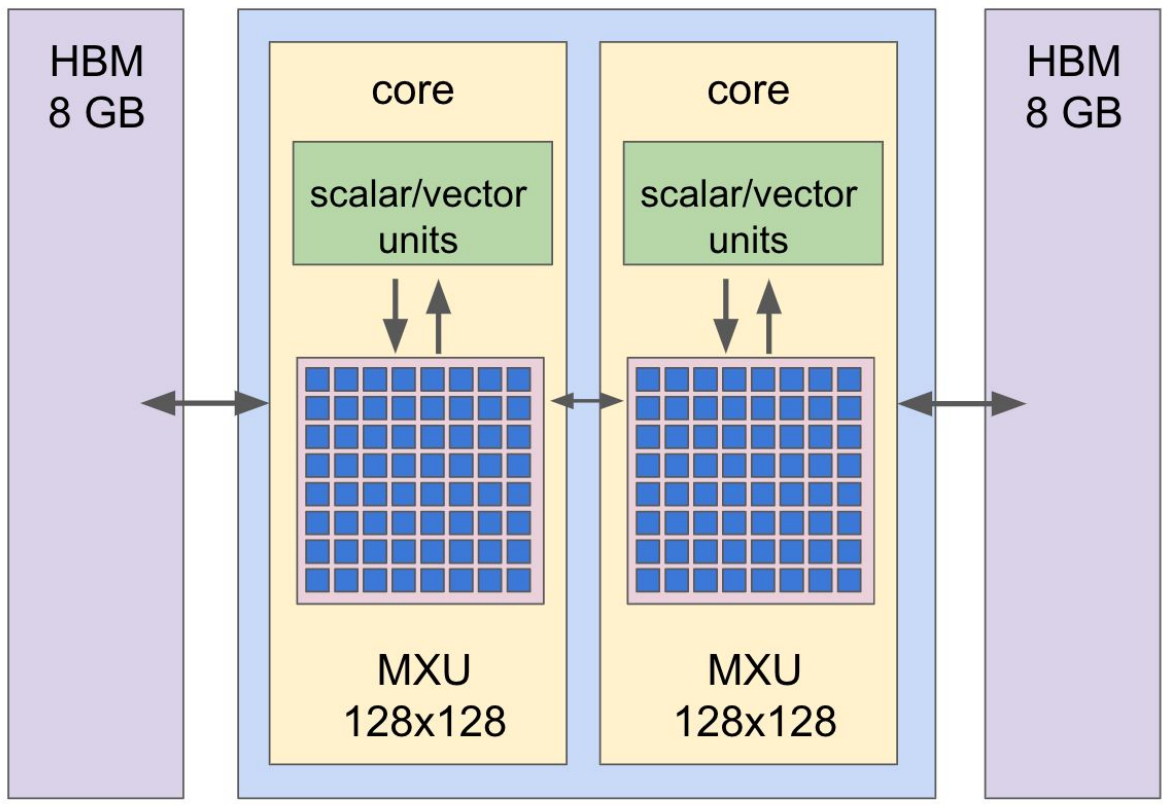

Figure 5: A block diagram of Google's Tensor Processing Unit v2 (TPUv2)

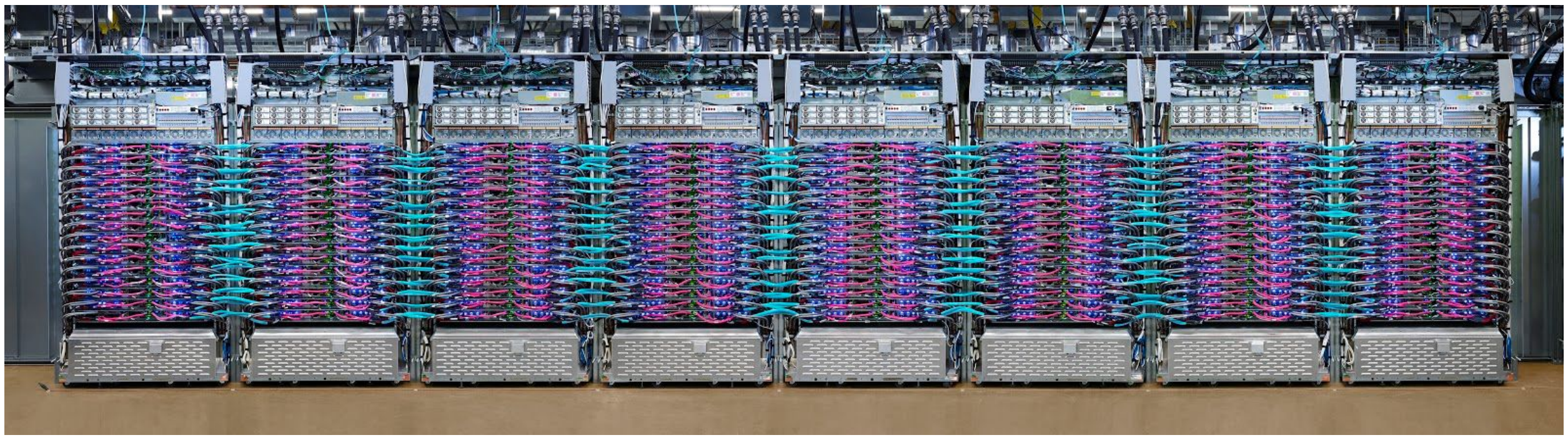

Figure 6: Google's TPUv3 Pod, consisting of 1024 TPUv3 chips w/peak performance of >100 petaflop/s

## Low Precision Numeric Formats for Machine Learning

TPUv2 and TPUv3 use a custom-designed floating point format called bfloat16 [Wang and Kanwar 2019], which departs from the IEEE half-precision 16-bit format to provide a format that is more useful for machine learning and also enables much cheaper multiplier circuits. bfloat16 was originally developed as a lossy compression technique to help reduce bandwidth requirements during network communications of machine learning weights and activations in the DistBelief system, and was described briefly in section 5.5 of the TensorFlow white paper [Abadi *et al.* 2016, sec. 5.5]. It has been the workhorse floating format in TPUv2 and TPUv3 since 2015. As of December, 2018, Intel announced plans to add bfloat16 support to future generations of Intel processors [Morgan 2018].

Figure 7 below shows the split between sign, exponent, and mantissa bits for the IEEE fp32 single-precision floating point format, the IEEE fp16 half-precision floating point format, and the bfloat16 format.

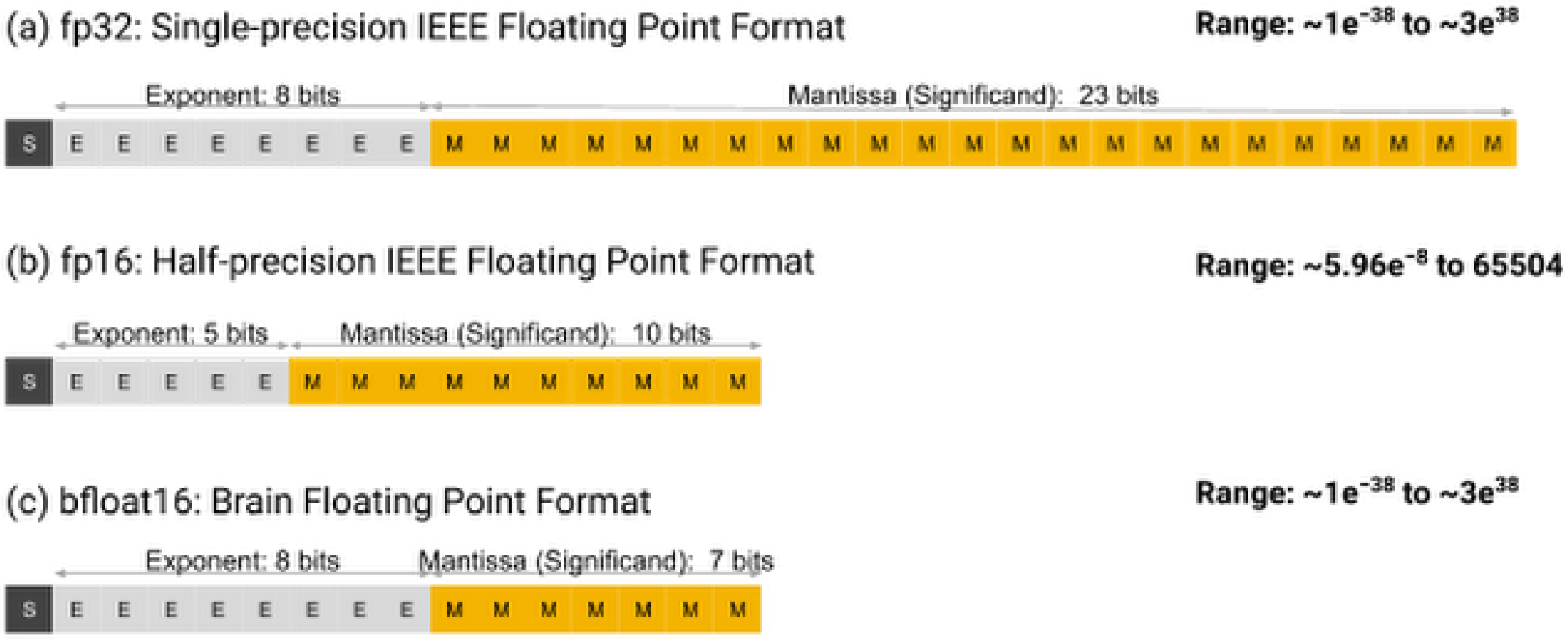

Figure 7: Differences between single-precision IEEE/half-precision IEEE/brain16 Floating Point Formats

As it turns out, machine learning computations used in deep learning models care more about dynamic range than they do about precision. Furthermore, one major area & power cost of multiplier circuits for a floating point format with *M* mantissa bits is the ($M$+1) $\times$ ($M$+1) array of full adders (that are needed for multiplying together the mantissa portions of the two input numbers. The IEEE fp32, IEEE fp16 and bfloat16 formats need 576 full adders, 121 full adders, and 64 full adders, respectively. Because multipliers for the bfloat16 format require so much less circuitry, it is possible to put more multipliers in the same chip area and power budget, thereby meaning that ML accelerators employing this format can have higher flops/sec and flops/Watt, all other things being equal. Reduced precision representations also reduce the bandwidth and energy required to move data to and from memory or to send it across interconnect fabrics, giving further efficiency gains.

### The Challenge of Uncertainty in a Fast Moving Field

One challenge for building machine learning accelerator hardware is that the ML research field is moving extremely fast (as witnessed by the growth and absolute number of research papers published per year shown in Figure 4). Chip design projects that are started today often take 18 months to 24 months to finish the design, fabricate the semiconductor parts and get them back and install them into a production datacenter environment. For these parts to be economically viable, they typically must have lifetimes of at least three years. So, the challenge for computer architects building ML hardware is to predict where the fast moving field of machine learning will be in the 2 to 5 year time frame. Our experience is that bringing together computer architects, higher-level software system builders and machine learning researchers to discuss co-design-related topics like “what might be possible in the hardware in that time frame?” and “what interesting research trends are starting to appear and what would be their implications for ML hardware?” is a useful way to try to ensure that we design and build useful hardware to accelerate ML research and production uses of ML.

## Machine Learning for Chip Design

One area that has significant potential is the use of machine learning to learn to automatically generate high quality solutions for a number of different NP-hard optimization problems that exist in the overall workflow for designing custom ASICs. For example, currently placement and routing for complex ASIC designs takes large teams of human placement experts to iteratively refine from high-level placement to detailed placement as the overall design of an ASIC is fleshed out. Because there is considerable human involvement in the placement process, it is inconceivable to consider radically different layouts without dramatically affecting the schedule of a chip project once the initial high level design is done. However, placement and routing is a problem that is amenable to the sorts of reinforcement learning approaches that were successful in solving games, like AlphaGo. In placement and routing, a sequence of placement and routing decisions all combine to affect a set of overall metrics like chip area, timing, and wire length. By having a reinforcement learning algorithm learn to “play” the game of placement and routing, either in general across many different ASIC designs, or for a particular ASIC design, with a reward function that combines the various attributes into a single numerical reward function, and by applying significant amounts of machine-learning computation (in the form of ML accelerators), it may be possible to have a system that can do placement and routing more rapidly and more effectively than a team of human experts working with existing electronic design tools for placement and routing. We have been exploring these approaches internally at Google and have early preliminary-but-promising looking results. The automated ML based system also enables rapid design space exploration, as the reward function can be easily adjusted to optimize for different trade-offs in target optimization metrics.

Furthermore, it may even be possible to train a machine learning system to make a whole series of decisions from high-level synthesis down to actual low-level logic representations and then perform placement and routing of these low-level circuits into a physical realization of the actual high level design in a much more automated and end-to-end fashion. If this could happen, then it's possible that the time for a complex ASIC design could be reduced substantially, from many months down to weeks. This would significantly alter the tradeoffs involved in deciding when it made sense to design custom chips, because the current high level of non-recurring engineering expenses often mean that custom chips or circuits are designed only for the highest volume and highest value applications.

# Machine Learning for Semiconductor Manufacturing Problems

With the dramatic improvements in computer vision over the past decade, there are a number of problems in the domain of visual inspection of wafers during the semiconductor manufacturing process that may be amenable to more automation, or to improved accuracy over the existing approaches in this area. By detecting defects earlier or more accurately, we may be able to achieve higher yields or reduced costs. A survey of these approaches provides a general sense of the area [Huang and Pan 2015]

# Machine Learning for Learned Heuristics in Computer Systems

Another opportunity for machine learning is in the use of learned heuristics in computer systems such as compilers, operating systems, file systems, networking stacks, etc. Computer systems are filled with hand-written heuristics that have to work in the general case. For example, compilers must make decisions about which routines to inline, which instruction sequences to choose which of many possible loop nesting structures to use, and how to lay out data structures in memory [Aho *et al.* 1986]. Low-level networking software stacks must make decisions about when to increase or decrease the TCP window size, when to retransmit packets that might have been dropped, and whether and how to compress data across network links with different characteristics. Operating systems must choose which blocks to evict from their buffer cache, which processes and threads to schedule next, and which data to prefetch from disk [Tanenbaum and Woodhull 1997]. Database systems choose execution plans for high-level queries, make decisions about how to lay out high level data on disks, and which compression methods to use for which pieces of data [Silberschatz *et al.* 1997].

The potential exists to use machine-learned heuristics to replace hand-coded heuristics, with the ability for these ML heuristics to take into account much more contextual information than is possible in hand-written heuristics, allowing them to adapt more readily to the actual usage patterns of a system, rather than being constructed for the average case. Other uses of ML can replace traditional data structures like B-trees, hash tables, and Bloom filters with learned index structures, that can take advantage of the actual distribution of data being processed by a system to produce indices that are higher performance while being 20X to 100X smaller [Kraska *et al.* 2018].

# Future Machine Learning Directions

A few interesting threads of research are occuring in the ML research community at the moment that will likely be even more interesting if combined together.

First, work on sparsely-activated models, such as the sparsely-gated mixture of experts model [Shazeer *et al.* 2017], shows how to build very large capacity models where just a portion of the model is "activated" for any given example (say, just 2 or 3 experts out of 2048 experts). The routing function in such models is trained simultaneously and jointly with the different experts, so that the routing function learns which experts are good at which sorts of examples, and the experts simultaneously learn to specialize for the characteristics of the stream of examples to which they are given. This is in contrast with most ML models today where the whole model is activated for every example. Table 4 in Shazeer *et al.* 2017 showed that such an approach be simultaneously ~9X more efficient for training, ~2.5X more efficient for inference, and higher accuracy (+1 BLEU point for a language translation task).

Second, work on automated machine learning (AutoML), where techniques such as neural architecture search [Zoph and Le 2016, Pham *et al.* 2018] or evolutionary architectural search [Real *et al.* 2017, Gaier and Ha 2019] can automatically learn effective structures and other aspects of machine learning models or components in order to optimize accuracy for a given task. These approaches often involve running many automated experiments, each of which may involve significant amounts of computation.

Third, multi-task training at modest scales of a few to a few dozen related tasks, or transfer learning from a model trained on a large amount of data for a related task and then fine-tuned on a small amount of data for a new task, has been shown to be very effective in a wide variety of problems [Devlin *et al.* 2018]. So far, most use of multi-task machine learning is usually in the context of a single modality (e.g. all visual tasks, or all textual tasks) [Doersch and Zisserman 2017], although a few authors have considered multi-modality settings as well [Ruder 2017].

A particularly interesting research direction puts these three trends together, with a system running on large-scale ML accelerator hardware, with a goal of being able to train a model that can perform thousands or millions of tasks in a single model. Such a model might be made up of many different components of different structures, with the flow of data between examples being relatively dynamic on an example-by-example basis. The model might use techniques like the sparsely-gated mixture of experts and learned routing in order to have a very large capacity model [Shazeer *et al.* 2017], but where a given task or example only sparsely activates a small fraction of the total components in the system (and therefore keeps computational cost and power usage per training example or inference much lower). An interesting direction to explore would be to use dynamic and adaptive amounts of computation for different examples, so that "easy" examples use much less computation than "hard" examples (a relatively unusual property in the machine learning models of today). Figure 8 depicts such a system.

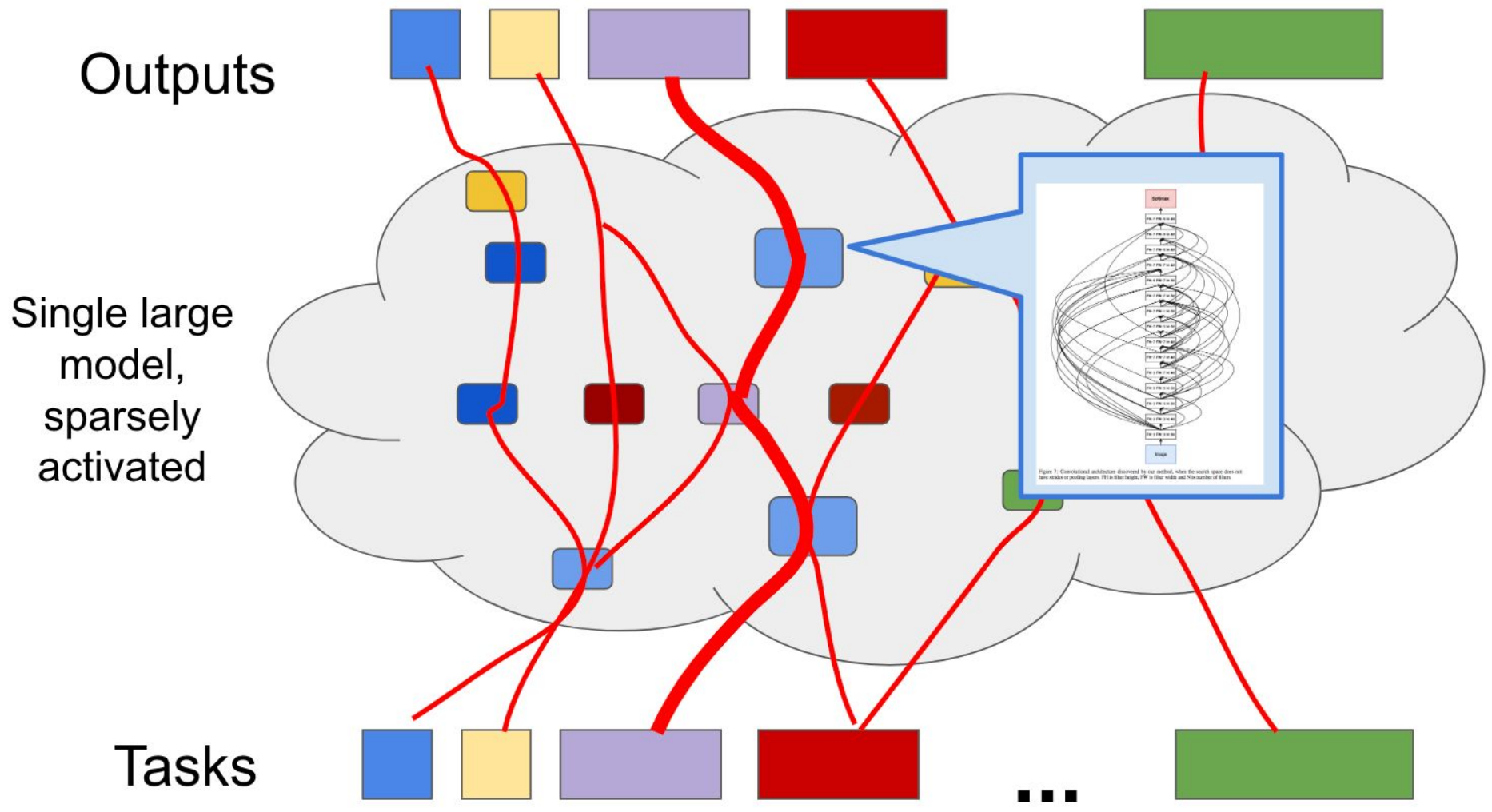

Figure 8: A diagram depicting a design for a large, sparsely activated, multi-task model. Each box in the model represents a component. Models for tasks develop by stitching together components, either using human-specified connection patterns, or automatically learned connectivity. Each component might be running a small architectural search to adapt to the kinds of data which is being routed to it, and routing decisions making components decide which downstream components are best suited for a particular task or example, based on observed behavior.

Each component might itself be running some AutoML-like architecture search [Pham *et al.* 2017], in order to adapt the structure of the component to the kinds of data that it is being routed to that component. New tasks can leverage components trained on other tasks when that is useful. The hope is that through very large scale multi-task learning, shared components, and learned routing, the model can very quickly learn to accomplish new tasks to a high level of accuracy, with relatively few examples for each new task (because the model is able to leverage the expertise and internal representations it has already developed in accomplishing other, related tasks).

Building a single machine learning system that can handle millions of tasks, and that can learn to successfully accomplish new tasks automatically, is a true grand challenge in the field of artificial intelligence and computer systems engineering: it will require expertise and advances in many areas, spanning solid-state circuit design, computer networking, ML-focused compilers, distributed systems, and machine learning algorithms in order to push the field of artificial intelligence forward by building a system that can generalize to solve new tasks independently across the full range of application areas of machine learning.

# Conclusion

The advances in machine learning over the past decade are already affecting a huge number of fields of science, engineering, and other forms of human endeavor, and this influence is only going to increase. The specialized computational needs of machine learning combined with the slowdown of general-purpose CPU performance improvements in the post-Moore's Law-era represent an exciting time for the computing hardware industry [Hennessy and Patterson 2019]: we now have a set of techniques that seem to be applicable to a vast array of problems across a huge number of domains, where we want to dramatically increase the scale of the models and datasets on which we can train these models, and where the impact of this work will touch a vast fraction of humanity. As we push the boundaries of what is possible with large-scale, massively multi-task learning systems that can generalize to new tasks, we will create tools to enable us to collectively accomplish more as societies and to advance humanity. We truly live in exciting times.

# Acknowledgements

Anand Babu, Alison Carroll, Satrajit Chatterjee, Jason Freidenfelds, Anna Goldie, Norm Jouppi, Azalia Mirhoseini, David Patterson, and Cliff Young, as well as this year's ISSCC chairs and anonymous ISSCC representatives all provided helpful feedback on the content of this article that was much appreciated.